\documentclass[11pt,a4paper]{article}

\usepackage{url}
\usepackage{times}
\usepackage{latexsym}
\usepackage{hyperref}

\usepackage{amsmath,amssymb}
\usepackage{graphicx}
\usepackage{tabularx}
\usepackage{multirow}
\usepackage{pifont}

\usepackage[toc,page]{appendix}
\usepackage{xcolor}
\usepackage{floatrow}

\newfloatcommand{capbtabbox}{table}[][\FBwidth]
\usepackage{arydshln}

\usepackage{minibox}

\usepackage{tabularx} %
\usepackage[noend]{algpseudocode}

\newcommand{\xxcomment}[4]{\textcolor{#1}{[$^{\textsc{#2}}_{\textsc{#3}}$ #4]}}

\newcommand{\ya}[1]{\xxcomment{red}{Y}{A}{#1}}

\newcommand{\eat}[1]{}

\newcommand{\nlstring}[1]{`#1'}

\newcommand{\ourmodel}{\text{Our Approach}}

\newcommand{\random}{\text{Random Interaction}}
\newcommand{\nointeract}{\text{No Interaction}}
\newcommand{\staticDT}{\text{No Initial Query Interaction}}

\renewcommand{\mid}{{\kern 0.075em}|{\kern 0.075em}}

\newcommand{\encoder}{\mathbf{enc}}

\usepackage{amsmath}

\newcommand{\X}{X}  %

\newcommand{\A}{\mathcal{A}}  %
\newcommand{\querysubset}{\mathcal{X}} %

\newcommand{\query}{x}

\newcommand{\step}[1]{#1}

\newcommand{\answer}{r}

\newcommand{\allanswers}{\mathcal{R}}
\newcommand{\goal}{y}

\newcommand{\rvgoal}{y}
\newcommand{\allgoals}{\mathcal{Y}}
\newcommand{\question}{q}

\newcommand{\allquestions}{\mathcal{Q}}

\usepackage{fancyhdr,graphicx,amsmath,amssymb}
\usepackage[ruled]{algorithm2e}

\usepackage{mathtools} 
\usepackage{esvect} 
\usepackage{amssymb}
\usepackage{bm}
\usepackage{bbm}
\usepackage{makecell}

\usepackage{floatrow}

\usepackage{blindtext}
\usepackage{graphicx}
\usepackage{booktabs}

\usepackage{subcaption}

\usepackage{titlesec}
\titlespacing*{\section} {0pt}{0.6ex plus 0.1ex minus .2ex}{0.6ex plus .2ex}
\titlespacing{\paragraph} {0pt}{0.2ex plus 0.1ex minus .2ex}{0.8em}

\usepackage[hyperref]{acl2020}
\usepackage{times}
\usepackage{latexsym}

\usepackage{microtype}

\aclfinalcopy %
\def\aclpaperid{2528} %

\title{Interactive Classification by  Asking Informative Questions}

\author{Lili Yu\textsuperscript{1}, Howard Chen\textsuperscript{1}, Sida I. Wang\textsuperscript{1,2},Tao Lei\textsuperscript{1} \hspace{-5pt} \and \hspace{-5pt} Yoav Artzi\textsuperscript{1,3}  \\
\textsuperscript{1}ASAPP Inc., New York, USA\\
\textsuperscript{2}Princeton University, New Jersey, USA \\
\textsuperscript{3}Cornell University, New York, USA \\
{\tt \{liliyu, hchen, tao\}@asapp.com}\\
{\tt sidaw@cs.princeton.edu} \ \hspace{0.2cm}
{\tt yoav@cs.cornell.edu}
}

\date{}

\begin{document}
\maketitle

\begin{abstract}

We study the potential for interaction in natural language classification. We add a limited form of interaction for intent classification, where users provide an initial query using natural language, and the system asks for additional information using binary or multi-choice questions. At each turn, our system decides between asking the most informative question or making the final classification prediction.The simplicity of the model allows for bootstrapping of the system without interaction data, instead relying on simple crowdsourcing tasks. We evaluate our approach on two domains, showing the benefit of interaction and the advantage of learning to balance between asking additional questions and making the final prediction.

\end{abstract}

\section{Introduction}
\label{sec:intro}

Responding to natural language queries through simple, single-step classification has been studied extensively in many applications,
including user intent prediction~\citep{csbot, intentprediction}, and information retrieval~\citep{kang2003query,rose2004understanding}. 
Typical methods rely on a single user input to produce an output, and do not interact with the user to reduce ambiguity and improve the final prediction. For example, users may under-specify a request due to incomplete understanding of the domain; or the system may fail to correctly interpret the nuances of the input query. In both cases, a low quality decision could be mitigated by further interaction with the user.

\begin{figure}[th]\centering
\includegraphics[width=1\linewidth]{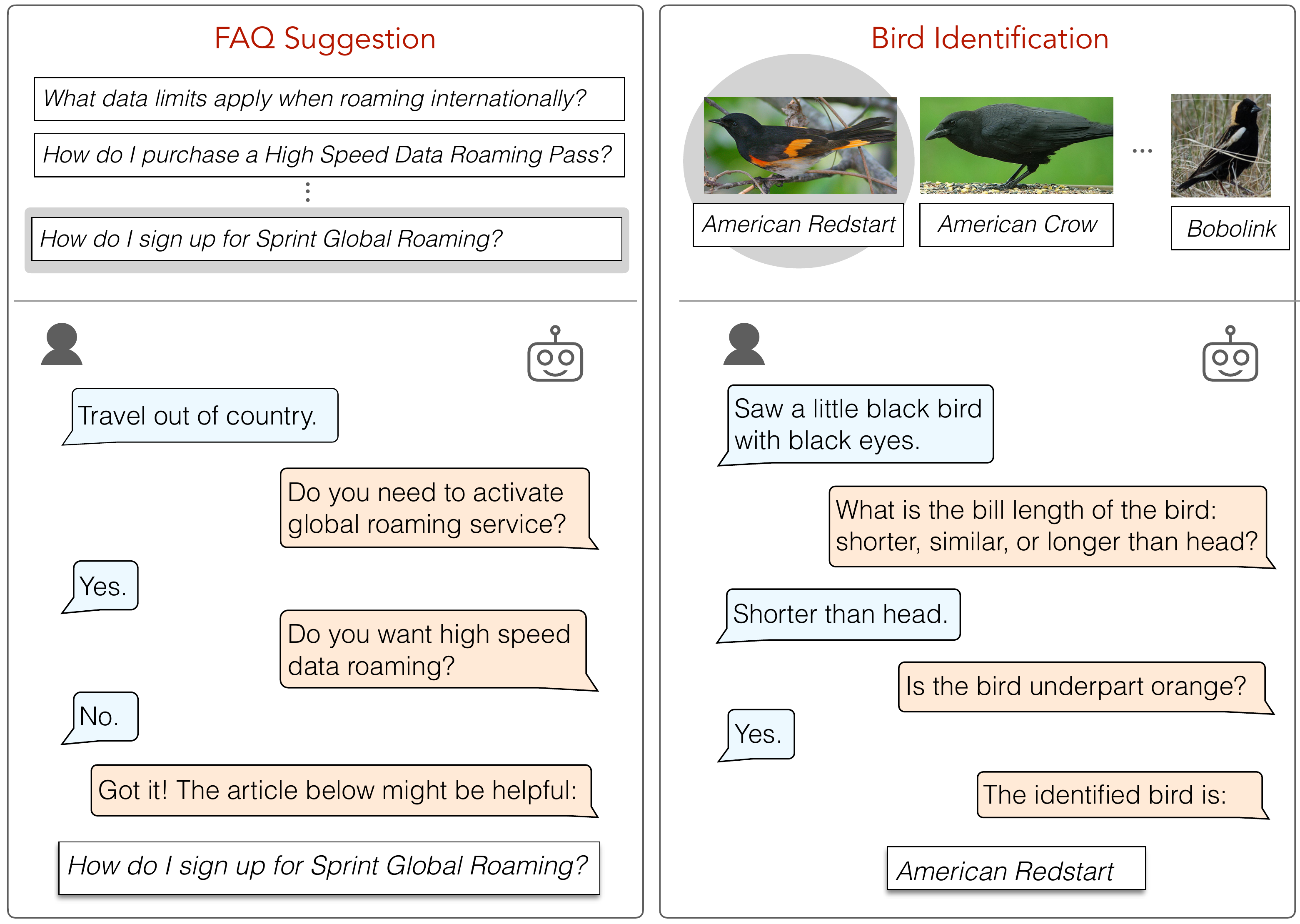}
\caption{Two examples of interactive classification systems: providing a trouble-shooting FAQ suggestion (left) and helping identifying bird species from a descriptive text query (right). The top parts show example classification labels: FAQ documents or bird species.\footnotemark The ground truth label of each interaction example is shaded. The lower parts show user interactions with the systems. The user starts with an initial natural language query. At each step, the system asks a clarification question. The interaction ends when the system returns an output label. 
}
\label{fig:example}
\vspace{1pt}
\end{figure}

In this paper we take a low-overhead approach to add limited interaction to intent classification. Our goal is two-fold: (a) study the effect of interaction on the system performance, and (b) avoid the cost and complexities of interactive data collection. 
We build an interactive system that poses a sequence of binary and multiple choice questions following the initial user natural language query. 
Figure~\ref{fig:example} illustrates  such interactions in two domains, showcasing the opportunity for clarification while avoiding much of the complexity involved in unrestricted natural language interactions.
\footnotetext{The images are for illustration only. Our approach does not use images.}
We design our approach not to rely on user interaction during learning, which requires users to handle low quality systems or costly Wizard of Oz experiments. 
We adopt a Bayesian decomposition of the posterior distributions over intent labels and user responses through the interaction process. 
We use the posteriors to compute question expected information gain, which allows us to efficiently select the next question at each interaction turn.
We balance between the potential increase in accuracy and the cost of asking additional questions with a learned policy controller that decides whether to ask additional questions or return the final prediction. 
We estimate each distribution in our posterior decomposition independently by crowdsourcing initial queries and keywords annotation. 
We use non-interactive annotation tasks that do not require Wizard-of-Oz style dialog annotations~\citep{Kelley:1984:IDM:357417.357420, 1604.04562}. During training, we train a shared text encoder to compare natural language queries, clarification questions, user answers and classification targets in the same embedding space. This enables us to bootstrap to unseen clarification targets and clarification questions, further alleviating the need of expensive annotation.

We evaluate our method on two public tasks: FAQ suggestion~\citep{Shah2018AdversarialDetection} and bird identification using the text and attribute annotations of the Caltech-UCSD Birds dataset~\citep{WahCUB_200_2011}.
The first task represents a virtual assistant application in a trouble-shooting domain, while the second task provides well-defined multiple-choice question annotations and naturally noisy language inputs.
We evaluate with both a simulator and human users. 
Our experiments show that adding user interaction significantly increases the classification accuracy. 
Given at most five turns of interaction, our approach improves the accuracy of a no-interaction baseline  by over 100\% on both tasks for simulated evaluation and over 90\% for human evaluation. 
Even a single clarification question provides significant accuracy improvements, 40\% for FAQ suggestion and 65\% for bird identification in our simulated analysis. 
Our code and data is available at \url{https://github.com/asappresearch/interactive-classification}.

\section{Technical Overview }
\label{sec:formulation}

Our goal is to classify a natural language query to a label through an interaction. 

\paragraph{Notation}

We treat the classification label $y$, interaction question $q$ and the user response $r$ as random variables.
We denote an assignment of a random variable using subscripts, such as $y = y_i$ and $q = q_j$. We use superscripts for the observed value of the random variable at a given time step, for example, $q^{\step{t}}$ is a question asked at time step $t$.
When clear from the context, we write $y_i$ instead of $y = y_i$.
For example, $p(r \mid q_j, y_i)$ denotes the conditional distribution of $r$ given $y = y_i$ and $q=q_j$, and $p(r_k \mid q_j, y_i)$ further specifies the corresponding probability when $r=r_k$.
 
An interaction starts with the user providing an initial user query $\query$. At each turn $t$, the system selects a question $\question^{\step{t}}$, to which the user responds with $\answer^{\step{t}}$, or returns a label $y$ to conclude the interaction.
We consider two types of questions: binary and multiple choice questions.
The predefined set of possible answers for a question $\question^{\step{t}}$ is $\allanswers(\question^{\step{t}})$, where $\allanswers(\question^{\step{t}})=\{ {\rm yes}, {\rm no} \}$ for binary questions, or a predefined set of question-specific values for multiple choice questions. 
We denote an interaction up to time $t$ as $\X^{\step{t}} = (\query, \langle(\question^{\step{1}}, \answer^{\step{1}}),\dots,(\question^{\step{t}}, \answer^{\step{t}}\rangle)$, and 
the set of possible class labels as $\allgoals = \left \lbrace \goal_1, \dots, \goal_N \right \rbrace$.
Figure~\ref{fig:example} shows example interactions in our two evaluation domains.

\paragraph{Model}
We model the interactive process using a parameterized distribution over class labels that is conditioned on the observed interaction (Section~\ref{sec:retrieval_prob}), a question selection criterion (Section~\ref{sec:mig}), and a parameterized policy controller (Section~\ref{sec:policy}).
At each time step $t$, we compute the belief of each $\goal_i \in \allgoals$ conditioned on $\X^{t-1}$.
The trained policy controller decides between two actions: to return the current best possible label or to obtain additional information by asking a question. 
The model selects the question with the maximal information gain.
Given a user response, the model updates the belief over the classification labels. 

\paragraph{Learning}
We use crowdsourced data to bootstrap model learning. The crowdsourcing data collection includes two non-interactive tasks. First, we obtain a set of user initial queries $\querysubset_i$ for each label $\goal_i$. For example, for an FAQ, \nlstring{How do I sign up for Spring Global Roaming}, an annotated potential initial query is \nlstring{Travel out of country}. Second, we ask annotators to assign text tags to each $\goal_i$, and heuristically convert these tags into a set of question-answer pairs $\A_i = \{ (\question_m, \answer_m) \}_{m=1}^{M_i} $, where $\question_m$ denotes a templated question and $\answer_m$ denotes the answer. For example, the question \nlstring{What is your phone operating system?} can pair with one of the following answers: \nlstring{IOS}, \nlstring{Android operating system},  \nlstring{Windows operating system} or \nlstring{Not applicable}. 
We denote this dataset as $\{ ( \goal_i, \querysubset_i, \A_i ) \}_{i=1}^N$.
We describe the data collection process in Section~\ref{sec:data}. 
We use this data to train our text embedding model (Section~\ref{sec:learn_retrieval}), to create a user simulator (Section~\ref{sec:method:sim}), and to train the policy controller  (Section~\ref{sec:policy}).

\paragraph{Evaluation}

We report classification the model accuracy, and study the trade-off between accuracy and the number of  turns that the system takes. We evaluate with both a user simulator and real human users. When performing human evaluation, we additionally collect qualitative ratings.

\section{Related Work}
\label{sec:related}

Human feedback has been leveraged to train natural language processing models, including for dialogue~\citep{LiDIALOGUEHUMAN-IN-THE-LOOP}, semantic parsing~\citep{Artzi:11, 1606.02447, IyerLearningFeedback} and text classification~\citep{DBLP:journals/corr/abs-1805-03818}. 
These methods collect user feedback after the model-predicting stage and treat user feedback as additional offline training data to improve the model. In contrast, our model leverages  user interaction to increase  prediction performance. Human feedback has been incorporated in reinforcement learning as well, for example to learn a reward function from language as reflecting human preferences~\citep{rlhuman}.

Language-based interaction has been studied in the context of visual question answering ~\citep{DeVriesGuessWhatDialogue,LeeAnswererDialogue,ChattopadhyayEvaluatingGames,DasLearningLearning, 1902.08355, MIGRL}, SQL generation~\citep{P18-1124, 1910.05389}, information retrieval~\citep{SpokenContentRetrieval,clarificatonIR} and multi-turn text-based question answering~\citep{rao2018,ReddyCoQA:Challenge,quac}.
Most methods require learning from recorded dialogues~\citep{Wu18:q20rinna,2018arXiv180807645H,LeeAnswererDialogue,rao2018} or conducting Wizard-of-Oz dialog annotations~\citep{Kelley:1984:IDM:357417.357420, 1604.04562}.
Instead, we limit the interaction to multiple-choice and binary questions. 
This simplification allows us to reduce the complexity of data annotation while still achieving effective interaction.
Our task can be viewed as an instance of the popular 20-question game (20Q), which has been applied to a celebrities knowledge base~\citep{chen2018learning,2018arXiv180807645H}. Our approach differs in using  natural language descriptions of classification targets, questions and answers to compute our distributions, instead of treating them as categorical or structural data.

Our question selection method is  related to several existing methods. 
\citet{kovashka2013attribute} refine image search by asking to compare visual qualities against selected reference images, and \citet{LeeAnswererDialogue} perform object identification in an image by posing binary questions about the object or its location. 
Both methods, as well as ours use an entropy reduction criterion to select the best next question. 
We use a Bayesian decomposition of the joint distribution, which can be easily extended to other model-driven selection methods.
\citet{rao2018} propose a learning-to-ask approach by modeling the expected utility of asking question. 
Our selection method can be considered as a special case when entropy is used as the utility.
In contrast to \citet{rao2018}, we model the entire interaction history instead of a single turn of follow-up questioning.
Our model is trained using crowdsourced annotations, while \citet{rao2018} uses real user-user interaction data. 
Alternatively to asking questions, \citet{ferecatu2007interactive} and \citet{1805.00145} present to the user the most likely image in an image retrieval scenario. The user compares it with the ground-truth image and provides feedback using relevance score or natural language describing the discrepancy between them.

\section{Method}
\label{sec:method}
We maintain a probability distribution $p(\goal\mid\X^{\step{t}})$ over the set of labels $\allgoals$. 
At each interaction step, we first update this belief, decide if to ask a question or return the classification output using a policy controller and, if needed, select a question to ask using information gain. 

\subsection{Belief Probability Decomposition}
\label{sec:retrieval_prob}

We decompose the conditional probability \mbox{$p(\goal = \goal_i\mid\X^{\step{t}} )$} using Bayes rule:
\begin{align*}
p(\goal_i \mid\X^{\step{t}})=\; & p(\goal_i\mid\X^{\step{t-1}},\question^{\step{t}}, \answer^{\step{t}}) \\
\propto \; & p(\answer^{\step{t}},\question^{\step{t}},\goal_i\mid\X^{\step{t-1}}) \\
=  \;&  p(\question^{\step{t}}\mid\goal_i,\X^{\step{t-1}}) \ p(\goal_i\mid\X^{\step{t-1}})\\
& ~~~~~~ p(\answer^{\step{t}}\mid\question^{\step{t}},\goal_i,\X^{\step{t-1}})\;. 
\end{align*}
We make  two simplifying assumptions as modeling choices. 
First, the user response depends only on the question $\question^{\step{t}}$ and the underlying target label $y_i$, and is independent of past interactions. 
While this independence assumption is unlikely to reflect the course of interactions, it allows to simplify $p(\answer^{\step{t}}\mid \question^{\step{t}}, y_i, \X^{\step{t-1}})$ to $p(\answer^{\step{t}}\mid \question^{\step{t}}, y_i)$.
Second, the selection of the next question  $\question^{\step{t}}$ is deterministic given the interaction history $\X^{\step{t-1}}$.
Therefore, $p(q = \question^{\step{t}}\mid y_i, \X^{\step{t-1}}) = 1$, or zero for $q\neq \question^{\step{t}}$. Section~\ref{sec:mig} describes this process. 
We rewrite the decomposition as:
\begin{equation}
\label{eq:progupdate}
\begin{split}
p(\goal_i \mid\X^{\step{t}})\ \propto p(\answer^{\step{t}}\mid \question^{\step{t}},\goal_i)\cdot 1 \cdot p(\goal_i\mid\X^{\step{t-1}}) \\
= \, p(\goal_i \mid \query) \ \prod_{\tau=1}^{\step{t}} \; p(\answer^{\step{\tau}} \mid \question^{\step{\tau}},\goal_i)\;.
\end{split}
\end{equation}
Predicting the classification label given the observed interaction $\X^{\step{t}}$ is reduced to modeling $p(\goal_i\mid \query)$ and $p(\answer_k\mid \question_j, \goal_i)$, the label $\goal_i$ probability given the initial query $\query$ only and the probability of user response $\answer_k$ conditioned on the chosen question $\question_j$ and class label $\goal_i$.
This factorization enables leveraging separate annotations to learn the two components directly, alleviating the need for collecting costly recordings of user interactions.

\subsection{Information Gain Question Selection}
\label{sec:mig}

The system selects the question $\question^{\step{t}}$ to ask at turn $t$ to maximize the efficiency of the interaction.
We use a maximum information gain criterion.
Given $\X^{\step{t-1}}$, we compute the information gain on classification label $\rvgoal$ as the decrease on entropy by observing possible answers to question $\question$:
\begin{align*}
IG(\rvgoal\, ; \question\mid\X^{\step{t-1}})=H(\rvgoal\mid\X^{\step{t-1}}) - H(\rvgoal\mid\X^{\step{t-1}},\question)\;,
\end{align*}
where $H(\cdot\mid\cdot)$ denotes the conditional entropy.
Intuitively, the information gain measures the amount of information obtained about the variable $\goal$ by observing the value of another variable $\question$. 
Because the first entropy term $H(\rvgoal \mid \X^{\step{t-1}})$ is a constant regardless of the choice of $\question$, the selection of $\question^{\step{t}}$ is equivalent to $\question^{\step{t}} = \arg\min_{\question_j} H(\rvgoal\mid\X^{\step{t-1}},\question_j)$, where
\begin{align*}
H(\rvgoal\mid\X^{\step{t-1}},\question_j) =& \sum_{ \answer_k \in \allanswers(\question_j)} p(\answer_k \mid \X^{\step{t-1}},\question_j) \\
& \qquad~~   H(\rvgoal\mid\X^{\step{t-1}}, \question_j, \answer_k) \\
H(\rvgoal\mid\X^{\step{t-1}}, \question_j, \answer_k) =&  \sum_{\goal_i \in \allgoals} p(\goal_i\mid\X^{\step{t-1}}, \question_j,\answer_k)\ \\  
&  \qquad \log p(\goal_i\mid\X^{\step{t-1}}, \question_j,\answer_k) \\
p(\answer_k\mid\X^{\step{t-1}},\question_j) =& \sum_{\goal_i\in\allgoals} p(\answer_k, \goal_i \mid \X^{\step{t-1}}, \question_j) \\
  =& \sum_{\goal_i\in\allgoals} p(\answer_k\mid\question_j,\goal_i ) \\ & \qquad p(\goal_i \mid \X^{\step{t-1}})\;.
\end{align*}
\noindent
We use the independence assumption (Section~\ref{sec:retrieval_prob}) to calculate $p(\answer_k\mid\X^{\step{t-1}},\question_j)$. Both $p(\answer_k\mid\X^{\step{t-1}},\question_j)$ and $p(\goal_i\mid\X^{\step{t-1}}, \question_j,\answer_k)$ can be iteratively updated using $p(\goal_i\mid \query)$ and $p(\answer_k\mid \question_j, \goal_i)$ as the interaction progresses (Equation~\ref{eq:progupdate}) to efficiently compute the information gain.

\subsection{Modeling the Distributions}
\label{sec:learn_retrieval}

We model $p(\goal_i \mid \query)$ and $p(\answer_k \mid \question_j, \goal_i)$ by encoding the natural language descriptions of questions, answers and classification labels. In our domains, the text representation of a label is the FAQ document or the bird name. 
We do not simply treat the labels, questions and answers as categorical variables.
Instead, we leverage their natural language content to estimate their correlation This reduces the need for heavy annotation and improves our model in low-resource scenarios. 
We use a shared neural encoder $\mathbf{enc}(\cdot)$ parameterized by $\psi$ to encode all texts.
Both probability distributions are computed using the dot-product score: $S(u, v) = \encoder(u)^\top\encoder(v)$, where $u$ and $v$ are two pieces of text. 
The probability of predicting the label $\goal_i$ given an initial query $\query$ is: 
\begin{equation*}
p(\goal_i\mid\query) = \frac{\exp(S(\goal_i, \query))}{\sum_{\goal_j \in \allgoals}\exp(S(\goal_j, \query))}\;.
\end{equation*}
The probability of an answer $\answer_k$ given a question $\question_j$ and label $\goal_i$ is a linear combination of the observed empirical distribution $\hat{p}(\answer_k\mid\question_j,\goal_i)$ and a parameterized estimation $\tilde{p}(\answer_k\mid\question_j,\goal_i)$: 
\begin{equation*}
    p(\answer_k\mid\question_j,\goal_i) =  \lambda\hat{p}(\answer_k\mid\question_j,\goal_i) + (1-\lambda)\tilde{p}(\answer_k\mid\question_j,\goal_i)\;,
\end{equation*}
where $\lambda\in [0,1]$ is a hyper-parameter.
We use the question-answer annotations $\A_i$ for each label $\goal_i$ to estimate $\hat{p}(\answer_k\mid\question_j,\goal_i)$ using  empirical counts. 
For example, in the FAQ suggestion task, we collect multiple user responses for each question and class label, and average across annotators to estimate $\hat{p}$ (Section~\ref{sec:data}). 
The second term $\tilde{p}(\answer_k\mid\question_j,\goal_i)$ is computed using the text encoder:
\begin{align*}
&\tilde{p}(\answer_k\mid\question_j,\goal_i) \\ & ~~~~~ = \frac{\exp(w \cdot S(\question_j\#\answer_k, \goal_i)+b)}{\sum_{\answer_l \in \allanswers(\question_j)} \! \exp(w \cdot S(\question_j\#\answer_l, \goal_i)+b)}\;,
\end{align*}
where $w, b\in\mathbb{R}$ are scalar parameters and $\question_j\#\answer_k$ is a concatenation of the question $\question_j$ and the answer $\answer_k$.\footnote{For example, for a templated question \nlstring{What is your phone operating system?} and an answer \nlstring{IOS}, $\question_m$ = \nlstring{phone operating system} and $\answer_m$ = \nlstring{IOS}, therefore, $\question_m\#\answer_m$ = \nlstring{phone operating system IOS}.}
Because we do not collect complete annotations to cover every label-question pair, $\tilde{p}$ provides a smoothing of the partially observed counts using the learned encoding $S(\cdot)$.

We estimate the parameters $\psi$ of $\encoder(\cdot)$ by pre-training using a dataset $\{ ( \goal_i, \querysubset_i, \A_i ) \}_{i=1}^N$, where $\goal_i$ is a label, $\querysubset_i$ is a set of initial queries and $\A_i$ is a set of question-answer pairs.
We create from this data a set of text pairs $(u,v)$ to train the scoring function $S(\cdot)$. For each label $\goal_i$, we create pairs $(\query, \goal_i)$ for each initial query $\query \in \querysubset_i$. We also create $(\question_m\#\answer_m,\goal_i)$ for each question-answer pair $(\question_m, \answer_m) \in \A_i$. 
We minimize the cross-entropy loss using  gradient descent:
\begin{equation*}
\mathcal{L}(\psi) = 
- S(u, v) + \log\sum_{v'}\exp(S(u, v'))\;.
\end{equation*}
The second term requires summation over all $v'$, which are all the labels in $\allgoals$. We approximate this sum using  negative sampling that replaces the full set $\allgoals$ with a sampled subset in each training batch.
The parameters $\psi$, $w$ and $b$ are fine-tuned using reinforcement learning during training of the policy controller (Section~\ref{sec:policy}).

\subsection{User Simulator}
\label{sec:method:sim}

We use a held-out dataset to build a simple simulator. We use the simulator to train the policy controller (Section~\ref{sec:policy}) and for performance analysis, in addition to human evaluation. 
The user simulator provides initial queries to the system and responds to the system initiated clarification questions.
The dataset includes $N$ examples  $\{ ( \goal_i, \querysubset'_i, \A_i') \}_{i=1}^N$, where  $\goal_i$ is a goal, $\querysubset'_i$ is a set of initial queries and $\A_i' = \{ (\question_m, \answer_m) \}_{m=1}^{M'_i}$ is a set of question-answer pairs.
While this data is identical in form to our training data, we keep it separated from the data used to estimate $S(\cdot)$, $p(\goal_i \mid \query)$ and $p(\answer_k \mid \question_j, \goal_i)$  (Section~\ref{sec:learn_retrieval}).
We estimate the simulator question response distribution $p'(\answer_k \mid \question_j, \goal_i)$ using smoothed empirical counts from the data.

At the beginning of a simulated interaction, we sample a target label $\hat{\goal}$, and sample a query $\query$ from the associated query set $\querysubset'$ to start the interaction.
Given a system clarification question $\question^{\step{t}}$ at turn $t$, the simulator responds with an answer $\answer^{\step{t}} \in \allanswers(\question^{\step{t}})$ by sampling from  $p'(\answer\mid\question^{\step{t}},\hat{\goal})$. Sampling provides natural noise to the interaction, and our model has no knowledge of $p'$. 
The interaction ends when the system returns a label, which we can then evaluate, for example to compute a reward in Section~\ref{sec:policy}. This setup is flexible in that the user simulator can be easily replaced or extended by a real human, and the system can be further trained with a human-in-the-loop setup.

\begin{algorithm}[t]
\small
\caption{Training procedure}
Estimate $p(\goal \mid \query )$ and $p(\answer \mid \question, \goal)$ with $w$ and $b$ randomly initialized \\
Estimate $p'(\answer \mid \question, \goal)$ for the user simulator \\
\For{{\rm episode} = 1 \dots M}{
Sample $(\query, \hat{\goal})$ from dataset\\
    \For{t = 1 \dots T} {
        Compute $p(\goal\mid\X^{\step{t-1}})$ (Equation~\ref{eq:progupdate})\\
        ${\rm action} = f(p(\goal\mid\X^{\step{t-1}}),t - 1;\theta)$ \\
        
        \uIf{${\rm action}$ is \textit{STOP}} {
            break \\} 
        \uElseIf{${\rm action}$ is \textit{ASK}}
        {
            $\question^{\step{t}} = \arg\max_{\question_j\in\allquestions}\ {\rm IG}(\rvgoal\,;\,\question_j\mid\X^{\step{t-1}})$\\
            $\answer^{\step{t}} \sim p'(\answer\mid\question^{\step{t}},\hat{\goal})$\\
        }
    }
$\goal^* = \arg\max_{\goal_i}\ p(\goal_i \mid \X^{\step{t-1}})$ \\
Compute the return (i.e., total reward) for every step $t$ using $\goal^*$ and $\hat{\goal}$\\
Update $w$, $b$, $\theta$ using policy gradient
}
\label{alg:algo_disjdecomp}
\end{algorithm}

\subsection{Policy Controller}
\label{sec:policy}

The policy controller decides at each turn $t$ to either select another question to query the user or to conclude the interaction. This provides a trade-off between exploration by asking questions and exploitation by returning the most probable classification label. The policy controller $f(\cdot, \cdot;\theta)$ is a feed-forward network parameterized by $\theta$ that takes the top-$k$ probability values and current turn $t$ as input. It generates one of two actions: \textit{STOP} or \textit{ASK}. 
When selecting \textit{ASK}, a question is selected to maximize the information gain. For \textit{STOP}, the label $\goal_i$ with highest probability is returned using $\arg\max_{\goal_i \in \allgoals} p(\goal_i \mid \X^{\step{t-1}})$ and the interaction ends.

\subsection{Training Procedure}

Algorithm~\ref{alg:algo_disjdecomp} describes the complete training process. 
First, we estimate  $p(\goal \mid \query )$ and $p(\answer \mid \question, \goal)$. We use randomly initialized and fixed  $w$ and $b$ parameters. 
We also estimate $p'(\answer \mid \question, \goal)$ for the user simulator (Section~\ref{sec:method:sim}). 
We then learn the policy controller using the user simulator with a policy gradient method. We use the REINFORCE algorithm~\citep{williams1992simple}.   The reward function  provides a positive reward for predicting the correct target at the end of the interaction, a negative reward for predicting the wrong target, and a small negative reward for every question asked. 
We learn the policy controller $f(\cdot, \cdot;\theta)$, and estimate $w$ and $b$ in  $p(\answer_k\mid\question_j,\goal_i)$ by back-propagating through the policy gradient. We keep the $\encoder(\cdot)$ parameters fixed during policy gradient.

\section{Data Collection}
\label{sec:data}
We design a crowdsourcing process to collect data for the FAQ task using Amazon Mechanical Turk.\footnote{https://www.mturk.com/} For the Birds domain, we re-purpose an existing dataset. 
We collect initial queries and tags for each FAQ document. 
Appendix~\ref{appendix:data} describes the worker training process. 

\paragraph{Initial Query Collection} 
We ask workers to consider the scenario of searching for an FAQ  document using an interactive system.
Given a target FAQ, we ask for an initial query that they would provide to such a system.
The set of initial queries that is collected for each document $\goal_i$ is $\querysubset_i$.
We encourage workers to provide incomplete information and avoid writing a simple paraphrase of the FAQ.
This process provides realistic and diverse utterances because users have limited knowledge of the system and the domain.

\paragraph{Tag Collection}
We collect natural language tag annotations for the FAQ documents. 
First, we use domain experts to define the set of possible free-form tags. The tags are not restricted to a pre-defined ontology and can be a phrase or a single word describing the topic of the document. 
We remove duplicate tags to finalize the set. 
Experts combine some binary tags to categorical tags. For example, tags \nlstring{IOS}, \nlstring{Android operating system} and \nlstring{Windows operating system} are combined to the categorical tag \nlstring{phone operating system}. 
We use a small set of deterministic, heuristically-designed templates to convert tags into questions. For example, the tag \nlstring{international roaming} is converted into a binary question \nlstring{Is it about international roaming?}; the categorical tag \nlstring{phone operating system} is converted into a multi-choice question \nlstring{What is your phone operating system?}.
Finally, we use non-experts to collect user responses to the questions by associating tags with FAQ targets. 
For binary questions, we ask workers to associate their tags to the FAQ target if they would respond \nlstring{yes} to the question. 
We show the workers a list of ten tags for a given target as well as a \nlstring{none of the above} option. 
Annotating all possible target-tag combinations is still expensive and most pairings are negative. We rank the tags based on the relevance against the target using $S(\cdot)$ trained only on the initial queries and show only the current top-50 to the workers. Later, we re-train $S(\cdot)$ on the complete data. 
For multi-choice questions, we show the workers a list of possible answers to a tag-generated question for a given FAQ. The workers need to choose one answer that they think best applies. They also have the option of choosing \nlstring{not applicable}. 
The workers do not engage in a multi-round interactive process. This allows for cheap and scalable collection.

\section{Experimental Setup}
\label{sec:exp}

\paragraph{Task I: FAQ Suggestion}
We use the FAQ dataset from \citet{Shah2018AdversarialDetection}. 
The dataset contains 517 troubleshooting documents from Sprint's technical website.
We collect 3,831 initial queries and 118,640 tag annotations using the setup described in Section~\ref{sec:data}. 
We split the data into 310/103/104 documents as training, development, and test sets.
Only the queries and tag annotations of the $310$ training documents are used for pre-training and learning the policy controller, leaving the queries and tag annotations in the development and test splits for evaluation only.

\vspace{-2pt}
\paragraph{Task II: Bird Identification}
We use the Caltech-UCSD Birds dataset~\citep[CUB-200;][]{WahCUB_200_2011}.
The dataset contains 11,788 bird images for 200 different bird species. 
Each bird image is annotated with a subset of 27 visual attributes and 312 attribute values pertaining to the color or shape of a particular part of the bird. 
We create categorical questions from attributes with less five possible values, providing eight categorical questions in total. The remaining 279 attributes are converted to binary questions.
Each image is annotated with 10 image captions describing the bird in the image~\citep{reed2016learning}. We use the image captions as initial user queries and bird species as labels. 
Since each caption contains only partial information about the bird species, the data is naturally noisy and provides challenging user interactions.
We do not use the images from the dataset for model training. The images are only provided for grounding during human evaluation.

\paragraph{Baselines}
We compare with four methods: 
\begin{itemize}\itemsep 0.5pt
\item {\nointeract:} the classification label is predicted using only the initial query. We consider four implementations: (1) BM25: a common keyword-based scoring model for retrieval methods~\citep{INR-019}; (2) $\text{RoBERTa}_{\text{BASE}}$: we use  a fine-tuned $\text{RoBERTa}_{\text{BASE}}$ model~\citep{roberta} as text encoder; (3) RNN: we use a recurrent neural network (RNN) with simple recurrent unit recurrence~\citep[SRU;][]{Lei:18sru} as text encoder, together with a fastText word embedding layer~\citep{fasttext}; and (4) RNN + self-attention: the same RNN neural model with a multi-head self-attention layer~\citep{self-attention, NIPS2017_7181}.
\item {\random:} at each turn, the system randomly selects a question to present the user. After $T$ turns, the classification label is chosen according to the belief $p(\goal\mid\X^{\step{T}})$. 
\item {\staticDT:} the system selects questions  without conditioning on the initial user query using maximum information criterion. This is equivalent to using a static decision tree to pick the question, always asking the same first question~\citep{Utgoff1989,Ling:2004:DTM:1015330.1015369}.
\item Variants of \ourmodel: we consider several variants of our full model. 
First, we replace the policy controller with two termination strategies: (1) end the interaction when $\max{p(\goal\mid\X^{\step{t}})}$ passes a threshold, or (2) end the interaction after a fixed number of turns.
Second, we disable the parameterized estimator $\tilde{p}(\answer_k\mid\question_j,\goal_i)$ by setting $\lambda=1$.
\end{itemize}

\paragraph{Evaluation}
We use human evaluation, and further analyze performance using our simulator. For human evaluation, users interact with our systems and baseline models using a web-based interactive interface. Each interaction starts with a user scenario:\footnote{Each scenario is related to a single groundtruth label and serves to ground user interactions.} a bird image or a device-troubleshooting scenario described in text.
The user types an initial query and answers follow-up questions selected by the system.
Once the system returns its prediction, we measure its accuracy, and the user is asked to rate the whole interaction according to rationality and naturalness.\footnote{We also surveyed users for  perceived correctness, but observed it is interpreted identically to rationality. Therefore, we omit this measure. }  The user does not know the correct target label. 
We use a five-points Likert score for the followup  questions. 
For FAQ Suggestion, we consider two evaluation setups: (1) assuming the model has access to tags in the development and test set for interaction, and (2) using only tags in the training set annotation. 
The former is equivalent to adding tags for new documents not seen during training time. The latter zero-shot evaluation setup allows us to investigate the model's performance on unseen targets with no additional tags associated with them. 
Appendix~\ref{appendix:human evaluation} provides further details of the human evaluation setup. 
We do further analysis with the user simulator . We evaluate classification performance using Accuracy@$k$, which is the percentage of time the correct target appears among the top-$k$ predictions of the model. 

\paragraph{Implementation Details}
We use the same encoder to encode initial queries, question-answer pairs and FAQ documents in the FAQ suggestion task. 
In the bird identification task, where the structure of bird names differs from the other texts, we use one encoder  for user initial queries and question-answer pairs and a second encoder for bird names. 
The policy controller receives a reward of 20 for returning the correct target label, a negative reward of -10 for the wrong target,  
and a turn penalty of -0.5 for each question asked. 
For our simulated analysis, we report the averaged results as well as the standard derivation from three independent runs for each model variant and baseline.
Appendix~\ref{appendix:implementation details} provides more implementation and training details.

\section{Results}
\label{sec:results}

Our simulated analysis shows that the SRU RNN text encoder performs better or similar to the other encoders. This encoder is also the most lightweight. Therefore, we use it for the majority of our experiments.

\begin{figure*}[t]
    \centering
    \includegraphics[width=1.0\textwidth,height=70pt]{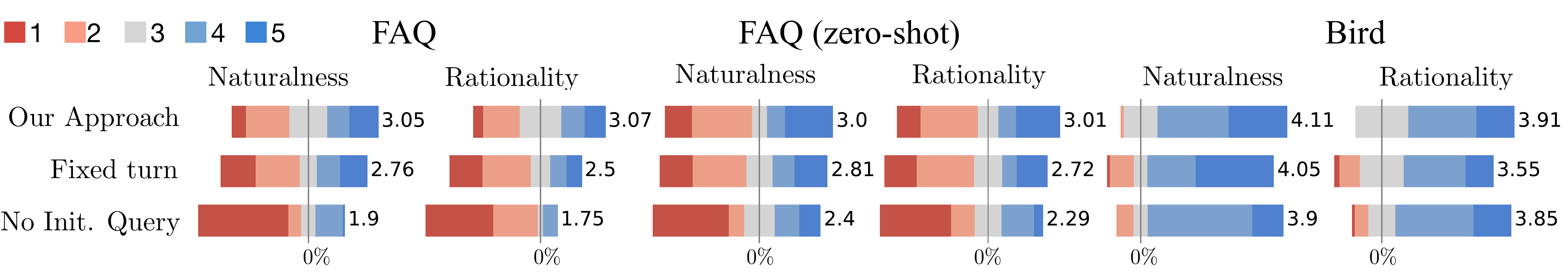}
    \caption{Human evaluation Gantt charts showing user ratings. We show the mean rating for each measure and system on the right of each bar. }
    \label{fig:humaneval}
\end{figure*}

\paragraph{Human Evaluation}
Figure~\ref{fig:humaneval} and Table~\ref{tab: humaneval-acc}  show the human evaluation results of our full model and three baselines: our approach with a fixed number of turns (four for FAQ and five for Bird), our approach without access to the initial query (No Init. Query) and our approach without interaction (No Int. (RNN)). 
Naturalness and rationality measure the quality of the interaction, so we show the results of the user survey in Figure~\ref{fig:humaneval} only for interactive systems. Because we do not ask users to fill the end-of-interaction survey for the no interaction baseline, we simply compute its numbers following the first query when evaluating our full approach. 
Our approach balances between accuracy and the user-centric measures, including naturalness and rationality, achieving stronger performance across the board.
All three models improve the classification performance with the addition of interaction.
Qualitatively, the users rate our full approach better than the two other interaction variants. 
This demonstrates that our model handles effectively real user interaction despite being trained with only non-interactive data.
We include additional details in Appendix~\ref{appendix:human evaluation}.

\begin{table}[t]
\centering
\footnotesize
\centering
\begin{tabular}{p{1.9cm}|c|c|c}
\toprule
     & FAQ & FAQ (zero-shot) &  Bird \\
\midrule
\ourmodel & \textbf{57\%} & \textbf{52\%} & \textbf{45\%} \\
\ourmodel & \multirow{2}{*}{53\%} & \multirow{2}{*}{47\%} & \multirow{2}{*}{37\%}  \\
~~~~ w/fixed turn  & & \\
No Init. Query & 43\% &  41\%  & 28\% \\
No Int. (RNN) & 30\% &  26\%  & 20\% \\
\bottomrule
\end{tabular}
\caption{Human evaluation classification accuracy.}
\label{tab: humaneval-acc}
\end{table}

\paragraph{Analysis with Simulated Interactions}

Table~\ref{table:result_table} shows performance using the the user simulator. We use these results to evaluate different choices beyond what is possible with human studies.  
We observe interaction is critical; removing the ability to interact decreases performance significantly. 
The {\random} and the {\staticDT} baselines both barely improve the performance over the {\nointeract}  RNN baseline,
illustrating the importance of guiding the interaction and considering the initial query.  
Our full model achieves an Accuracy@1 of 79\% for FAQ Suggestion and 49\% for Bird Identification using less than five turns, outperforming the {\nointeract} RNN baseline by 41\% and 26\%. 
When having no access to questions and answers in the development and test set during evaluation, the full model performance drops only slightly to 75\%, highlighting the model's ability to  generalize to unseen tags.
The two baselines with alternative termination strategies underperform the full model, indicating the effectiveness of the policy controller. 
The relatively low performance of the $\lambda=1$ variant, which effectively has fewer probability components leveraging natural language than our full model, and \staticDT\ confirm the importance of the learned natural language embedding encoder. 
Appendix~\ref{appendix:retrieval module}  includes further details on how different text encoders impact performance.

\begin{table*}[th]
\centering
\vspace{-5pt}
\footnotesize
\begin{tabular}{p{4.8cm}|c|c|c|c}
\toprule
   & \multicolumn{2}{c|}{ FAQ Suggestion} & \multicolumn{2}{c}{ Bird Identification }    \\
\midrule
    & Acc@1  & Acc@3   & Acc@1  & Acc@3         \\
\midrule
\nointeract\, (BM25)                  & $ 26\% $ & $ 31\%$ & \rm{N.A.} & \rm{N.A.}  \\
\nointeract\, ($\text{RoBERTa}_{\text{BASE}}$)               & $ 30\pm 0.5\% $ & $ 45\pm 0.6\%$  & $ 17\pm 0.3\% $ & $ 29\pm 0.3\%$  \\
\nointeract\, (RNN)                & $ 38\pm 0.5\% $ & $ 61\pm 0.3\%$  & $ 23\pm 0.1\% $ & $ 41\pm 0.2\%$  \\
\nointeract\, (RNN + self-attention)  & $ 39\pm 0.5\% $ & $ 63\pm 0.4\%$  & $ 23\pm 0.1\% $ & $ 41\pm 0.1\%$  \\
\hline
\random             & $ 39\pm 0.3\% ~(38 \pm 0.1\%) $ & $ 62\pm 0.4\%  ~(63 \pm 0.2\%)$   & $ 25\pm 0.1\% $ & $ 44\pm 0.1\%$  \\
\staticDT            & $ 46\pm 0.5\% ~(46 \pm 0.1\%) $ & $ 66\pm 0.6\%  ~(67 \pm 0.3\%)$  & $ 29\pm 0.2\% $ & $ 50\pm 0.3\%$  \\
\hline
\ourmodel   & $\textbf{ 79} \pm \textbf{ 0.7\%} ~(75 \pm 0.4\% ) $ & $\textbf{ 86} \pm \textbf{0.8\%} (83 \pm 0.4\%)  $ & $\textbf{ 49}  \pm \textbf{0.3\%}$ & $\textbf{69}  \pm \textbf{0.5\%}$  \\
\hspace{5mm} w/ threshold   & $ 73 \pm 0.6 \% ~(69 \pm 0.6\%)  $  & $ 82\pm 0.7\%  ~(81 \pm 0.6\%) $  & $ 41\pm 0.3\% $ & $ 59\pm 0.4\% $  \\  
\hspace{5mm} w/ fixed turn    & $ 71\pm 1.0\% ~(68 \pm 0.4\%)  $ & $ 81\pm 0.9\%  ~(81 \pm 0.6\%)  $  & $ 39\pm 0.2\% $ & $ 56\pm 0.4\%$  \\  
\hspace{5mm} w/ $\lambda=1$     & $ 66\pm 0.8\% ~(64 \pm 0.2\%)  $ & $ 71\pm 1.0\% ~(73 \pm 0.2\%) $  & $ 40\pm 0.1\% $ & $ 56\pm 0.2\%$  \\
\hline
\end{tabular}
\caption{Performance with simulated interactions. We evaluate our approach and several baselines using Accuracy@\{1, 3\}. Best performance numbers are in bold. We report the averaged results as well as the standard deviations from three independent runs for each model variant and baseline. For FAQ Suggestion, in parentheses, we provide zero-shot results, where the system has access to tags only for training questions.}
\label{table:result_table}
\end{table*}

\begin{figure*}[t!]
\centering
\vspace{-1pt}
\begin{subfigure}{.47\textwidth}
  \centering
\includegraphics[width=2.7in]{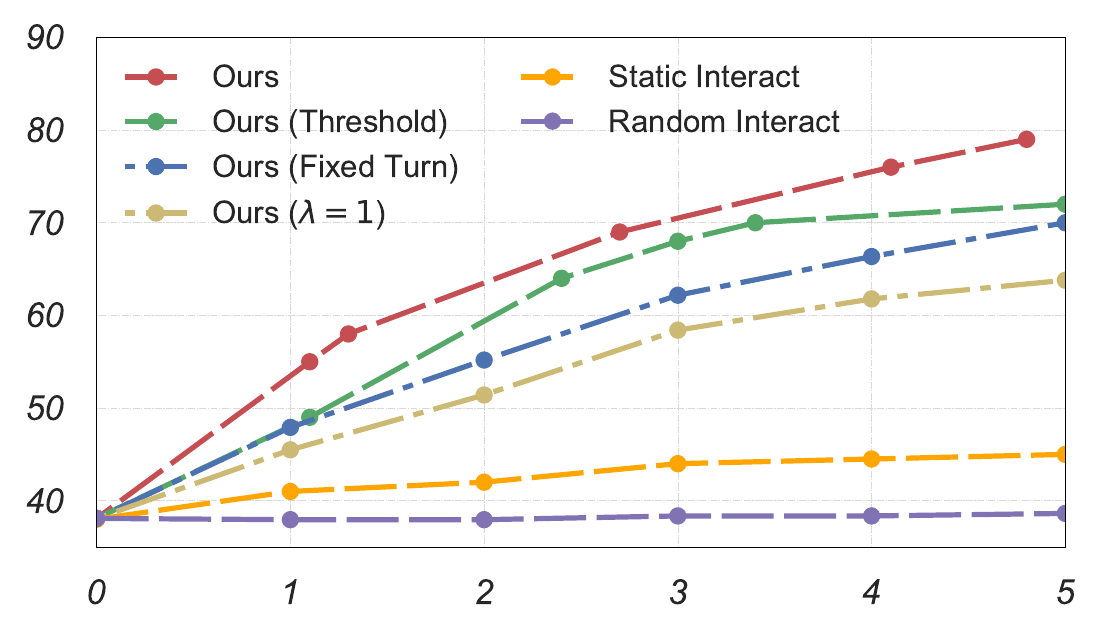}
\end{subfigure}%
\begin{subfigure}{.47\textwidth}
  \centering
\includegraphics[width=2.7in]{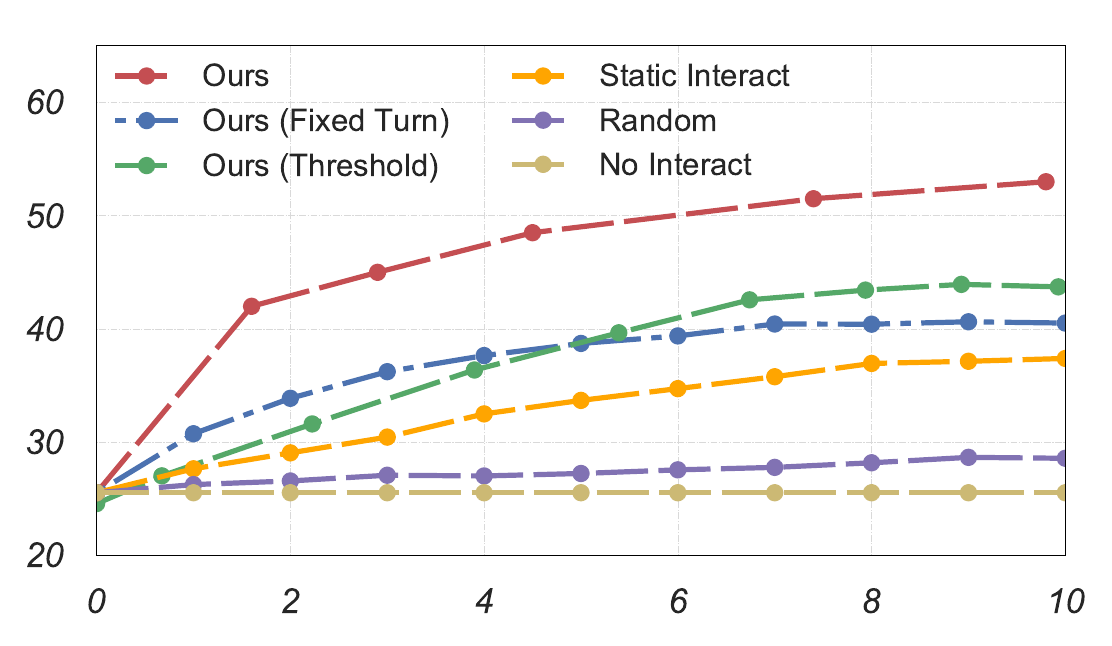}
\end{subfigure}
 \caption{Accuracy@1 (y-axis) against turns of interactions (x-axis) for FAQ (left) and Birds (right) tasks.}
\label{fig:result_fig}
\end{figure*}

\begin{figure*}[t!]
\centering
\includegraphics[width=1.0\linewidth]{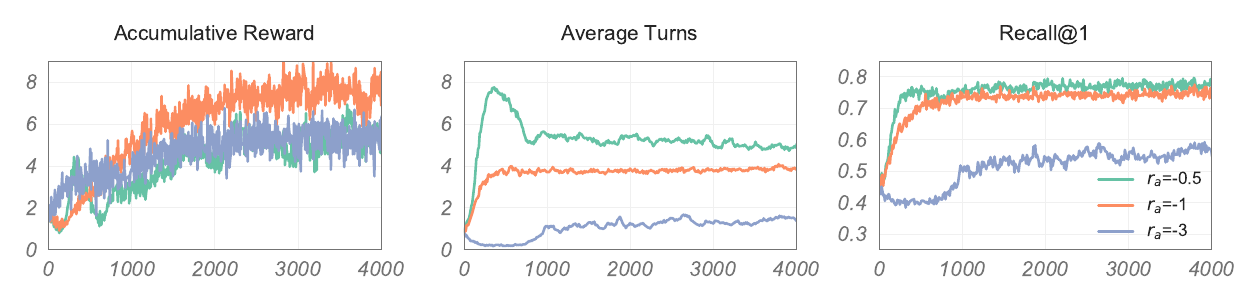}
\caption{Learning curves of our full model. We show accumulative reward (left), interaction turns (middle), and Accuracy@1 (right) on the test set, where x-axis is the number of episodes ($400$ trials per episode). The results are compared on different turn penalty $r_a$.}
\label{fig:RL training curve}
\end{figure*}

Figure~\ref{fig:result_fig} shows the trade-off between classification accuracy and the number of turns.
Each point on the plots is computed by varying the reward turn penalty for our model, the prediction threshold  and the predefined number of turns $T$.
Our model with the policy controller or the threshold strategy does not explicitly bound the number of turns, so we report the average number of turns across multiple runs for these two models. 
We achieve a relative accuracy boost of 40\% for FAQ and 65\% for Birds over no-interaction baselines with only one clarification question.
This highlights the value of leveraging human feedback to improve model accuracy in classification tasks. 

Figure~\ref{fig:RL training curve} shows the learning curves of our model with the policy controller trained with different turn penalties $r_a \in \{-0.5, -1, -3\}$.
We observe the models explore during the first 1,000 training episodes in the middle and the right plots.
The models achieve relatively stable accuracy after the early exploration stage.
The three runs end up using different numbers of expected turns because of the different $r_a$ values. 

\section{Conclusion}
\label{sec:disc}
We propose an approach for interactive classification, where the system can inquire missing information through a sequence of simple binary or multi-choice questions when users provide under-specified natural language queries. Our expert-guided, incremental design of questions and answers enables easy extension to add new classes, striking the balance between simplicity and extendability. Our modeling choices enable the system to perform zero-shot generalization to unseen classification targets and questions.
Our method uses information gain to select the best  question to ask at every turn, and a lightweight policy to efficiently control the interaction.
We demonstrate that the system can be bootstrapped without any interaction data and show effectiveness on two tasks.
A potential future research direction is to bridge the gap between this simple bootstrapping paradigm and the incorporation of user free-form responses to allow the system to handle free-text responses. 
We hope our work will encourage more research on different possibilities of building interactive systems that do not necessarily require handling  full-fledged dialogue, but still benefit from user interaction.

\section*{Acknowledgments}
We thank Derek Chen, Alex Lin, Nicholas Matthews, Jeremy Wohlwend, Yi Yang and the anonymous reviewers for providing valuable feedback on the paper. We would also like to thank Michael Griffths, Anna Folinsky and the ASAPP annotation team for their help on setting up and performing the human evaluation. Finally, we thank Hugh Perkins, Ivan Itzcovich, and Brendan Callahan for their support on the experimental environment setup.

\bibliography{anthology,9_refer}
\bibliographystyle{acl_natbib}

\appendix

\section{Appendices}
\label{sec:appendix}

\renewcommand\thefigure{\thesection.\arabic{figure}}    
\setcounter{figure}{0}

\renewcommand\thetable{\thesection.\arabic{table}} 
\setcounter{table}{0}

\subsection{Data collection}
\label{appendix:data}

We collect two types of data for the FAQ task. For the bird identification task we re-purpose existing data (Section~\ref{sec:exp}). 

\paragraph{Initial Query Collection Qualification} One main challenge for the data collection process is familiarizing the workers with the set of target documents. 
We set up a two-stage process to ensure the quality of the initial queries. The first stage is to write paraphrases of a given target, which is often a question in the FAQ task. 
We first allow the full pool of Amazon Mechanical Turk workers to perform the task. After that, we manually inspect the written queries and pick the ones that are good paraphrases of the FAQs. We selected $50$ workers that showed good understanding of the FAQs. 
In the second stage, workers are asked to provide initial queries with possibly insufficient information to identify the target. Out of the first 50 workers, we manually selected 25 based on the quality of the queries such as naturalness and whether they contain ambiguity or incompleteness by design. We used this pool of workers to collect 3,831 initial queries for our experiments.

\paragraph{Tag Association Qualification} 
The goal of this annotation task is to associate tags with classification labels. 
We train a model on the collected initial queries to rank tags for each classification target. 
We pick out the highest ranked tags as positives and the lowest ranked tags as negatives for each target. The worker sees in total ten tags without knowing which ones are the negatives. To pass the qualification task, the workers need to complete annotation on three targets without selecting any of the negative tags.

\begin{figure}[ht]
\centering
\includegraphics[width=3in]{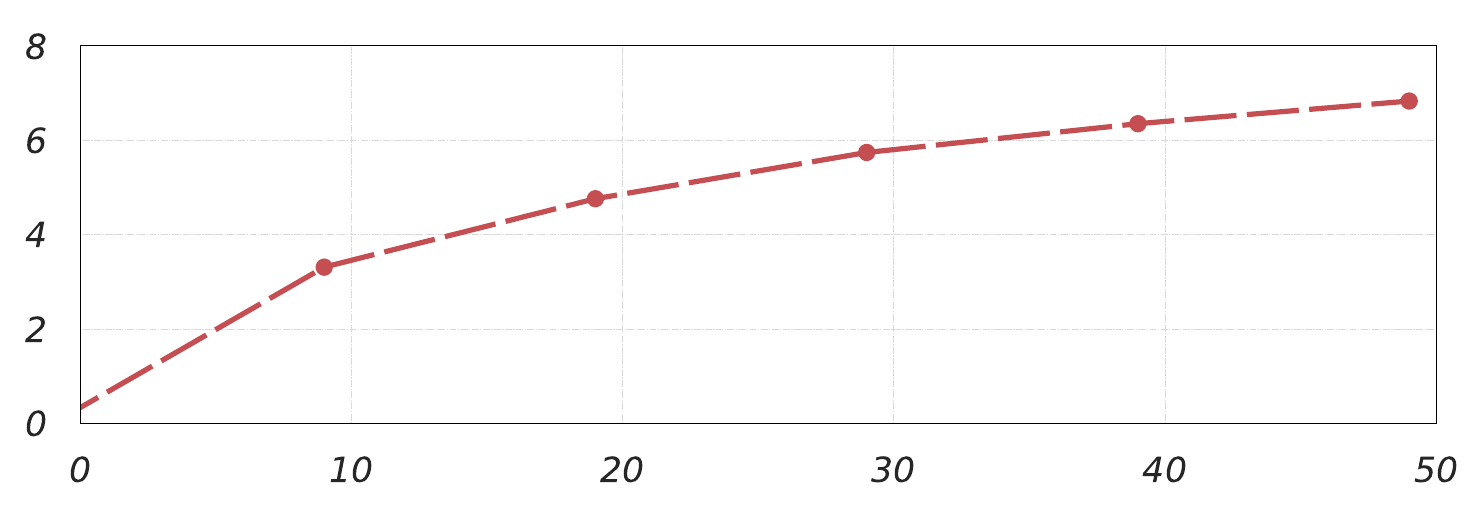}
\caption{Accumulated number of tags assigned to the targets (y-axis) by the workers against tag ranking (x-axis). The ranking indicates the relevance of the target-tag pairs from the pre-trained model. The curve plateaued at rank 50 suggesting that the lower ranked tags are less likely to be assigned to the target by the crowdsourcing workers.}
\label{fig:data_tag_collection}
\end{figure}

\paragraph{Tag Association Task Details}
After the qualification task, we take the top 50 possible tags for each target and split them into five non-overlapping lists (i.e., ten tags for each list) to show to the workers. 
Each of the lists is assigned to four separate workers to annotate.
We observe that  showing only the top-50 tags out of 813 is sufficient. Figure~\ref{fig:data_tag_collection} illustrates this: after showing the top-50 tags, the curve plateaus and no new tags are assigned to a target label. Table~\ref{table:user-simulator-stats} shows annotator agreement using Cohen's $\kappa$ score.

\begin{table}[ht]
\centering
\footnotesize
\begin{tabular}{p{1.55cm}||c|c|c|c|c}
\hline
\multicolumn{1}{c}{} & \multicolumn{5}{c}{\textbf{Tag Ranks}} \\
\multicolumn{1}{c}{} & \multicolumn{1}{c}{1-10} & \multicolumn{1}{c}{11-20} & \multicolumn{1}{c}{21-30} & \multicolumn{1}{c}{31-40} & \multicolumn{1}{c}{41-50} \\
\hline
Mean \# tags & 3.31 & 1.45 & 0.98 & 0.61 & 0.48\\
N.A. (\%) & 1.9 & 30.7 & 43.6 & 62.1 & 65.2\\
Mean $\kappa$ & 0.62 & 0.54 & 0.53 & 0.61 & 0.61 \\
\hline
\end{tabular}
\caption{Target-tag annotation statistics. We show five sets of tags to the annotators. The higher ranked ones are more likely to be related to the given target. The row mean \# tags is the mean number of tags that are annotated to a target, N.A. is the percentage of the tasks that are annotated as \textit{"none of the above"}, and mean $\kappa$ is the mean pairwise Cohen's $\kappa$ score.}
\label{table:user-simulator-stats}
\end{table}

\subsection{Implementation Details}
\label{appendix:implementation details}

We use a single-layer bidirectional Simple Recurrent Unit (SRU) as the encoder for the FAQ suggestion task and two layer bidirectional SRU for bird identification task. The encoder uses pre-trained fastText~\cite{fasttext} word embeddings of size 300, hidden size 150, batch size 200, and dropout rate 0.1.
The fastText embeddings remain fixed during training. 
We use the Noam learning rate scheduler \citep{NIPS2017_7181} with initial learning rate 1$e$-3, warm-up step 4,000 and a scaling factor of 2.0.
For the self-attention model, we use a multi-head self-attention layer with 16 heads and a hidden size of 64 for each head. The same dropout rate used for the text encoder is applied to the self-attention layer.
For the no interaction model with the \text{RoBERTa} encoder, we use the $\text{RoBERTa}_{\text{BASE}}$ model implemented by Hugghing Face~\citep{huggingface}. The $\text{RoBERTa}_{\text{BASE}}$ model is fine-tuned with learning rate of 1$e$-5, warmup step of 1,000, weight decay of 0.1, batch size of 16 and gradient accumulation step of 10. 
The policy controller is a two layer feed-forward network with a hidden layer of size 32 and ReLU activations.
The network takes the current turn and the top-$k$ values of the belief probabilities as input.
We choose $k=20$ and allow a maximum of 10 interaction turns.

\subsection{Additional Analysis}
\label{appendix:retrieval module}

We use the user simulator for further analysis of our system performance and alternative configurations.

\begin{table*}[ht]
    \small
    \centering
    \begin{tabular}{|l|l||c|c|c|c|c|c|c|c|}
    \multicolumn{2}{c}{\textbf{Text Input}} &
    \multicolumn{2}{c}{\textbf{Init Query}} &
    \multicolumn{2}{c}{\textbf{Init Query + Tags}} &
    \multicolumn{2}{c}{\textbf{Init + Paraphrase Query}} &
    \multicolumn{2}{c}{\textbf{Full Data}} \\
    \hline
    \multicolumn{1}{c}{init query} &     \multicolumn{1}{c|}{tags} &
    \multicolumn{1}{c}{Acc@1} & \multicolumn{1}{c||}{Acc@3} &
    \multicolumn{1}{c}{~~~Acc@1~~~} & \multicolumn{1}{c||}{Acc@3} &
    \multicolumn{1}{c}{~~~~~Acc@1~~~~~} & \multicolumn{1}{c||}{Acc@3} &
    \multicolumn{1}{c}{Acc@1} & 
    \multicolumn{1}{c}{Acc@3} \\
    \hline

    \multicolumn{1}{c}{\ding{51}} &
    \multicolumn{1}{c|}{\ding{55}} &
    \multicolumn{1}{c}{$0.28$} &
    \multicolumn{1}{c||}{$0.47$} &
    \multicolumn{1}{c}{$0.32$} &
    \multicolumn{1}{c||}{$0.51$} &
    \multicolumn{1}{c}{$0.35$} &
    \multicolumn{1}{c||}{$0.60$} &
    \multicolumn{1}{c}{$0.38$} &
    \multicolumn{1}{c}{$0.61$}\\
    \multicolumn{1}{c}{\ding{55}} &
    \multicolumn{1}{c|}{\ding{51}} &
    \multicolumn{1}{c}{$0.31$} & \multicolumn{1}{c||}{$0.50$} &
    \multicolumn{1}{c}{$0.57$} &
    \multicolumn{1}{c||}{$0.79$} &
    \multicolumn{1}{c}{$0.56$} &
    \multicolumn{1}{c||}{$0.74$} &
    \multicolumn{1}{c}{$0.70$} &
    \multicolumn{1}{c}{$0.87$}\\
    \multicolumn{1}{c}{\ding{51}} &
    \multicolumn{1}{c|}{\ding{51}} &
    \multicolumn{1}{c}{$0.36$} & \multicolumn{1}{c||}{$0.58$} &
    \multicolumn{1}{c}{$0.55$} &
    \multicolumn{1}{c||}{$0.79$} &
    \multicolumn{1}{c}{$0.63$} &
    \multicolumn{1}{c||}{$0.81$} &
    \multicolumn{1}{c}{$0.76$} &
    \multicolumn{1}{c}{$0.91$}\\
    \hline
    \end{tabular}
    \caption{Comparison of text encoders trained on different textual inputs and evaluated on three different prediction tasks. The model uses (a) the initial queries, (b) all attribute tags, or (c) both initial queries and tags as text inputs to predict the target. Each model is evaluated using Accuracy@\{1, 3\}.}
    \label{tab:pretrain-results}
\end{table*}

\begin{table*}[h]
\vspace{-5pt}
\small

\begin{tabular}{p{3.7cm}|c|c|c|c|c|c}
\toprule
   & \multicolumn{3}{c|}{ FAQ Suggestion } & \multicolumn{3}{c}{ Bird Identification}    \\
\midrule
    & RNN &  RNN + attention  &   $\text{RoBERTa}_{\text{BASE}}$   & RNN &  RNN + attention  &  $\text{RoBERTa}_{\text{BASE}}$     \\
\midrule
\nointeract            & $ 38\%$  & $ 39\%$ & $30\%$    & $ 23\%$ & $ 23\% $ & $ 17\%$  \\
\hline
\random             & $ 39\%$ & $ 38\% $ & $ 31\%$      & $ 25\%$ & $ 24\% $ & $ 17\%$  \\
\staticDT           & $ 46\%$  & $ 41\% $ & $ 37\%$     & $ 29\%$ & $ 29\% $ & $ 21\%$  \\
\hline
\ourmodel~w/ threshold  & $ 72\%$ & $ 73 \% $ & $ 54\%$  & $ 41\%$ & $ 38\% $ & $ 33\%$  \\  
\ourmodel~w/ fixed turn  & $ 71\%$  & $ 68\% $ & $ 47\%$  & $ 39\%$  & $ 37\% $ & $ 32\%$  \\  
\ourmodel~w/ $\lambda=1$   & $ 66\%$  & $ 67\% $ & $ 52\%$ & $ 40\%$  & $ 37\% $ & $ 32\%$  \\  
\hline
\end{tabular}
\caption{Accuracy@1 of our system with three different encoders. For all experiments, maximal number of turns is set to five.}
\label{tab:more_results}
\end{table*}

\paragraph{Text Encoder Training}
Table~\ref{tab:pretrain-results} shows the breakdown analysis of different ways to train the text encoder. We use initial queries as well as paraphrase queries to train the encoder, which has around 16K target-query examples.
To analyze the effectiveness of tags in addition to initial queries, we generate pseudo-queries by combining existing queries with sampled subset of tags from the targets. This augmentation strategy is useful to improve the classification performance. We also observe that using the set of tags instead of initial queries as text inputs for a specific target label improves  classification performance, indicating that the designed tags can capture the target label well. Finally, when we concatenate user initial queries and tags and use that as text input to the classifier, we achieve Accuracy@1 of 76\%. In our full model, we achieve 79\% with only querying about five tags. 

\paragraph{Performances of Different Encoders}
Table~\ref{tab:more_results} show  our system performance with different text encoders for both tasks.

\subsection{Human Evaluation}
\label{appendix:human evaluation}

Each interaction session starts with presenting  a user scenario (e.g., a bird image or a phone issue).
The user types an initial natural language query and  answers follow-up questions selected by the system.

\begin{figure*}[h]
\centering
\includegraphics[width=6.2in]{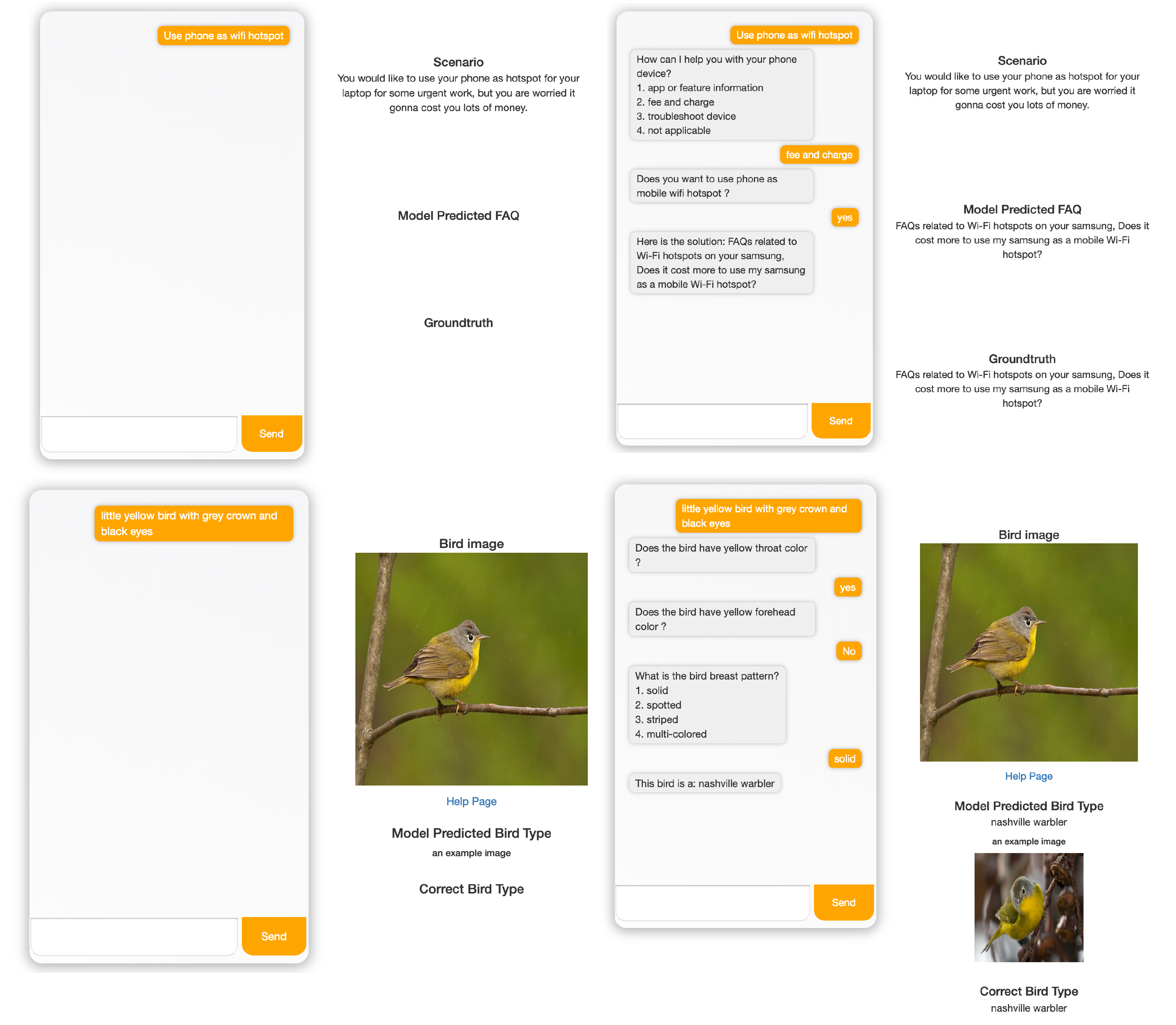}
\caption{The user interface for FAQ Suggestion  (top) and Bird Identification (bottom) tasks. The left panel shows the interface at the beginning of the interaction and the right panel shows the interface at the end of the interaction.}
\label{fig:ui}
\end{figure*}

\paragraph{FAQ Suggestion}
We design a user scenario for each target to present to the worker.
At the end of each interaction, the predicted FAQ and the ground truth are presented to the user, as shown in the top right panel in Figure~\ref{fig:ui}. The user answers the following questions: \nlstring{how natural is the interaction?} and \nlstring{do you feel understood by the system during the interactions?} on the scale of 1 (strongly disagree) to 5 (strongly agree), which we record as naturalness and rationality in Figure~\ref{fig:humaneval} and Table~\ref{tab: humaneval-acc}. 
Our full model performs best on Accuracy@1, naturalness and rationality. 
We show human evaluation examples in Table~\ref{table:faq_humaneval-examples}.

\begin{table*}[h]
\footnotesize
\begin{tabular}{ll}
\hline
\hline
  U: & How can I text my friends house phone?\\[0.15em]
  S: & Text message to landline? \\[0.15em]
  U: & Yes. \\[0.15em]
  S: & Do you want to send or receive text? \\[0.15em]
  U: & Yes. \\[0.15em]
  S: & Please read: \textit{How do I send a text message to landline?}\\[0.4em]
\hline
\hline
  U: & I'm in sprint's one up and want to upgrade my phone.\\[0.15em]
  S: & Do you want annual upgrade option? \\[0.15em]
  U: & Yes. \\[0.15em]
  S: & Do you want to exercise an option? \\[0.15em]
  U: & Yes. \\[0.15em]
  S: & Please read: \textit{I am enrolled in Sprint One Up. How do I exercise my annual upgrade option?}\\[0.4em]
\hline
\hline
  U: & Battery ran out immediately.\\[0.15em]
  S: & How can I help you with your phone device? \\[0.12em]
      &App or feature information; Fee and charge; Troubleshooting; Not applicable \\[0.15em]
  U: & Troubleshooting. \\[0.15em]
  S: & Are you using mobile wifi hotspot? \\[0.15em]
  U: & Yes. \\[0.15em]
  S: & Please read: \textit{Does using my phone as a mobile Wi-Fi hotspot drain my battery?}\\
\hline
\hline
\end{tabular}
\caption{Three examples of user interactions from the FAQ suggestion human evaluation.
}
\label{table:faq_humaneval-examples}
\end{table*}

\begin{figure*}[h]
\centering
\includegraphics[width=6.2in]{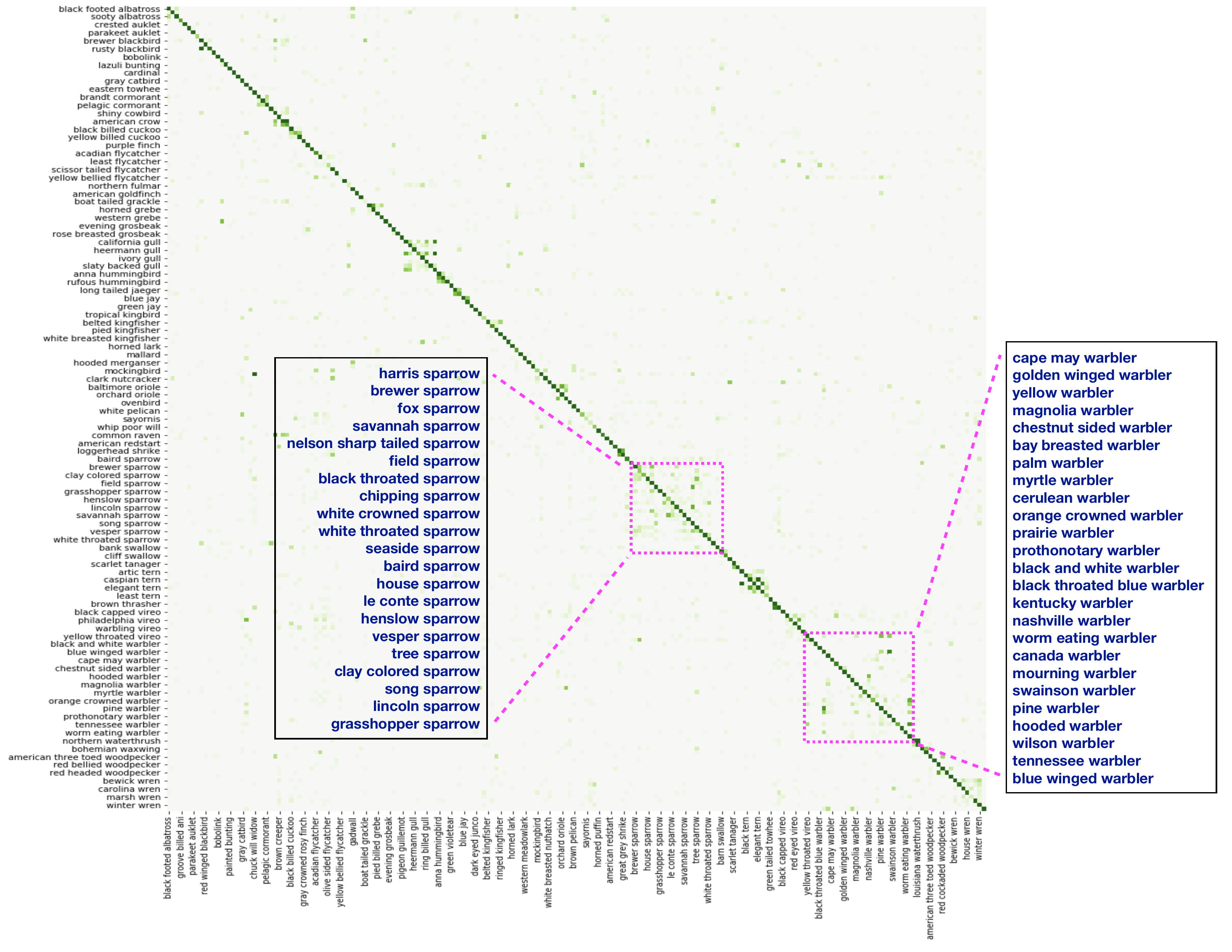}
\caption{Confusion matrix of our final output for bird identification task.}
\label{fig:bird_cm}
\end{figure*}

\begin{figure*}[h]
\centering
\includegraphics[width=6.2in]{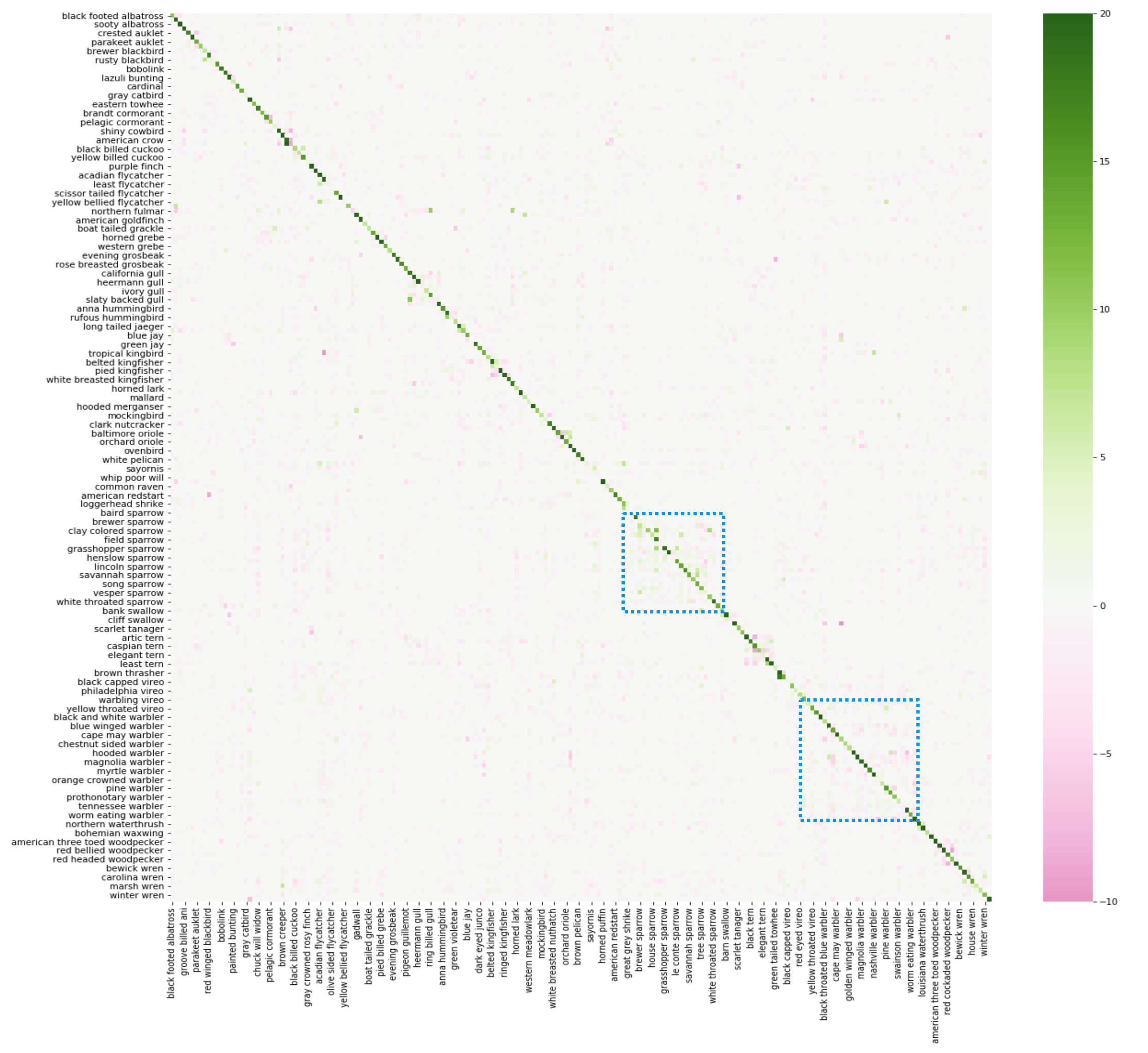}
\caption{Confusion matrix difference between the initial query with and without the interactions. High values along the diagonal and low values elsewhere are good.}
\label{fig:mc_diff}
\end{figure*}

\paragraph{Bird Identification}
The interface for bird identification task is similar to the FAQ suggestion task. Instead of presenting a scenario, we show a bird image to the user. The user needs to describe the bird to find out its category, which is analogous to writing an initial query. When answering system questions about attributes, we allow the user to reply \nlstring{not visible} if part of the bird is hidden or occluded. 
Given this reply, the system stops asking binary questions from the same label group. For example, if a user replies  \nlstring{not visible} to a the question \nlstring{does the bird has a black tail?}, then questions such as  \nlstring{does the bird has yellow tail?} and  \nlstring{does the bird has red tail?} will be skipped for the rest of the interaction. 
At the end of the interaction, the predicted and ground-truth bird images along with their categories are presented to the user as illustrated at the bottom right panel in Figure~\ref{fig:ui}. The user fills out a  questionnaire as in FAQ domain. The bird identification task is very challenging because of  its fine-grained categories, where many bird images look almost identical while belonging to different classes. 
Our full system improves classification accuracy from $20\%$ to $45\%$ against non-interactive baselines after less than three turns of interaction.
To better understand the task and the model behavior, we show the confusion matrix of the final model prediction after interaction in Figure~\ref{fig:bird_cm}. Of the 200 bird classes, there are 21 different kinds of sparrows and 25 different warbler. Most of the model errors are due to mistakes between these fine-grained bird classes. Figure~\ref{fig:mc_diff} shows how the confusion matrix changes when adding the interaction. The model makes improvement in distinct and also similar bird types.

\end{document}